\documentclass{article}

\usepackage[parfill]{parskip}
\usepackage{amsmath}
\usepackage{enumitem}
\usepackage{bm}
\usepackage{subcaption}
\usepackage{xcolor,colortbl}
\usepackage{caption}
\usepackage{array}
\usepackage{pdflscape}
\usepackage{multirow}
\usepackage{multicol}
\usepackage{rotating}
\usepackage{afterpage}

\usepackage{arxiv}
\usepackage[utf8]{inputenc} % allow utf-8 input
\usepackage[T1]{fontenc}    % use 8-bit T1 fonts
\usepackage{hyperref}       % hyperlinks
\usepackage{url}            % simple URL typesetting
\usepackage{booktabs}       % professional-quality tables
\usepackage{amsfonts}       % blackboard math symbols
\usepackage{nicefrac}       % compact symbols for 1/2, etc.
\usepackage{microtype}      % microtypography
\usepackage{lipsum}
\usepackage{graphicx}
\usepackage[round]{natbib}
\graphicspath{ {./images/} }

\title{Mitigating Spatial Disparity in Urban Prediction Using Residual-Aware Spatiotemporal Graph Neural Networks: A Chicago Case Study}

\author{
 Dingyi Zhuang \\
  Department of Civil and Environmental Engineering\\
  Massachusetts Institute of Technology\\
  Cambridge, MA 02139, USA \\
  \texttt{dingyi@mit.edu} \\
   \And
 Hanyong Xu \\
  Department of Urban Studies and Planning\\
  Massachusetts Institute of Technology\\
  Cambridge, MA 02139, USA \\
  \texttt{hanyongx@mit.edu} \\
  \And
 Xiaotong Guo \\
  Department of Civil and Environmental Engineering\\
  Massachusetts Institute of Technology\\
  Cambridge, MA 02139, USA \\
  \texttt{xtguo@mit.edu} \\
  \And
  Yunhan Zheng \\
  Singapore–MIT Alliance for Research \\ and Technology Centre (SMART) \\
  Singapore 138602 \\
  \texttt{yunhan@mit.edu} \\
  \And
  Shenhao Wang \\
  Department of Urban and Regional Planning \\
  University of Florida \\
  Gainesville, Florida, USA \\
  \texttt{shenhaowang@ufl.edu}\\
  \And
  Jinhua Zhao \\
  Department of Urban Studies and Planning\\
  Massachusetts Institute of Technology\\
  Cambridge, MA 02139, USA \\
  \texttt{jinhua@mit.edu}
}

\begin{document}
\maketitle
\begin{abstract}
Urban prediction tasks, such as forecasting traffic flow, temperature, and crime rates, are crucial for efficient urban planning and management. 
However, existing Spatiotemporal Graph Neural Networks (ST-GNNs) often rely solely on accuracy, overlooking spatial and demographic disparities in their predictions. 
This oversight can lead to imbalanced resource allocation and exacerbate existing inequities in urban areas. 
This study introduces a Residual-Aware Attention (RAA) Block and an equality-enhancing loss function to address these disparities. 
By adapting the adjacency matrix during training and incorporating spatial disparity metrics, our approach aims to reduce local segregation of residuals and errors. 
We applied our methodology to urban prediction tasks in Chicago, utilizing a travel demand dataset as an example.
%datasets including travel demand, crime, and accident reports.
Our model achieved a 48\% significant improvement in fairness metrics with only a 9\% increase in error metrics. 
Spatial analysis of residual distributions revealed that models with RAA Blocks produced more equitable prediction results, particularly by reducing errors clustered in central regions. 
Attention maps demonstrated the model's ability to dynamically adjust focus, leading to more balanced predictions. 
Case studies of various community areas in Chicago further illustrated the effectiveness of our approach in addressing spatial and demographic disparities, supporting more balanced and equitable urban planning and policy-making.
\end{abstract}

% keywords can be removed

\section{Introduction}

Urban prediction tasks involve predicting various urban indicators (e.g., traffic flow, temperature, etc.) using urban big data. 
These predictions are crucial for understanding urban patterns, which in turn benefit urban public administration and transportation management \citep{wang2024boosting}. 
Recent studies have applied various machine learning and deep learning techniques to improve prediction accuracies in downstream tasks such as travel demand forecasting, crime prediction, accident prediction, and weather forecasting \citep{Derrow_She_Wong_Lange_Hester_Perez_Nunkesser_Lee_Guo_Wiltshire_2021, Fu_Meng_Ye_Wang_2020,mehrabi_survey_2022}. 
These enhanced prediction models not only aid in more efficient resource allocation and policy-making but also play a vital role in addressing challenges related to urbanization, such as congestion, safety, and environmental sustainability.

Among these methods, Spatiotemporal Graph Neural Networks (ST-GNNs) have become prominent tools for urban prediction in various domains. 
ST-GNNs integrate the capabilities of Graph Neural Networks (GNNs) with architectures designed to process sequential temporal data. 
This combination makes ST-GNNs particularly effective at capturing both spatial and temporal dependencies in urban data.
Thus, ST-GNNs are able to model the intricate relationships between different urban regions and time points, providing significant advantages over traditional methods in delivering more accurate and reliable forecasts.

\begin{figure}[htbp]
\centering
\begin{subfigure}{.45\textwidth}
  \centering
  \includegraphics[width=.95\linewidth]{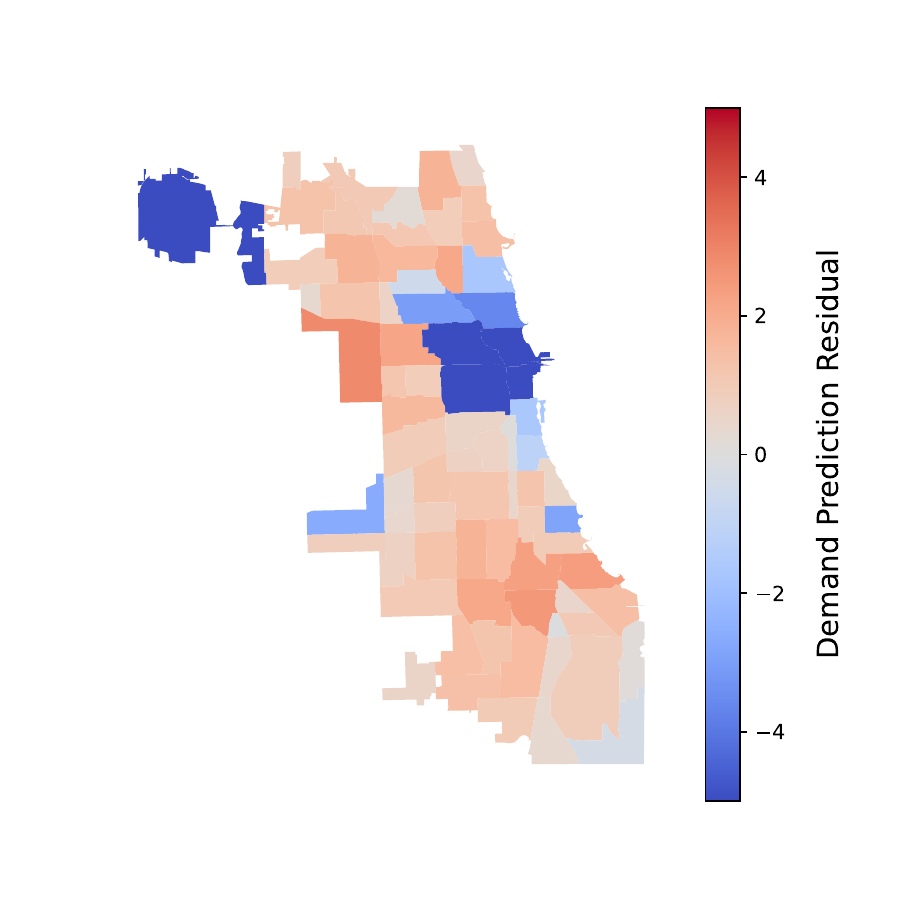}  
  \caption{STGCN Prediction Residual in Chicago.}
  \label{fig:vanilla_residual}
\end{subfigure}
\begin{subfigure}{.45\textwidth}
  \centering
\includegraphics[width=.95\linewidth]{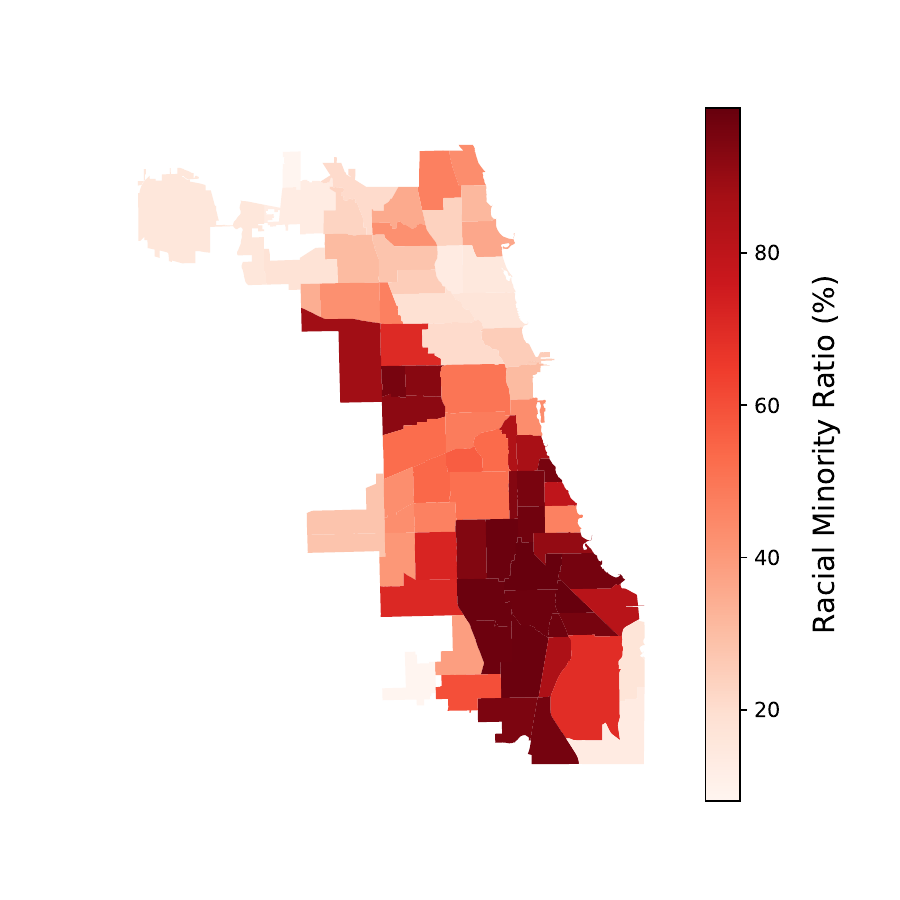}  
  \caption{Minority Rate in Chicago}
  \label{fig:minority_rate}
\end{subfigure}
\caption{Comparison between prediction residual and demographic distributions. (a): the red colors represent under-prediction while the blue colors represent over-prediction; (b): distribution of the racial minority groups in Chicago, defined as the percentage of the non-white population in each community area.}
\label{fig:res_compare_demo}
\end{figure}

A significant issue with previous ST-GNN models is their exclusive reliance on accuracy as the primary metric for evaluating model performance, without considering the potential social impact of the model's predictions. 
Figure \ref{fig:res_compare_demo} illustrates that the model STGCN's predictions are positively correlated with minority rates in Chicago \cite{Yu2018Spatio-TemporalForecasting}. 
This correlation is evidenced by positive residuals in regions with high minority rates and over-predictions in regions with low minority rates.
This discrepancy reveals distinct spatial patterns in over-prediction and under-prediction, highlighting the spatial disparities inherent in the model's predictions. 
These spatial disparities are intrinsically linked to demographic disparities and segregation phenomena. 
See Figure \ref{fig:minority_rate}, in the Chicago region, different racial groups predominantly reside in the northern and southern parts of the city, exacerbating the impact of demographic segregation on the model's performance.

There are two primary sources of disparity in ST-GNN model predictions: systematic sociodemographic biases in data collection and inherent model biases in ST-GNNs that propagate these data biases. 
Disparities stemming from data collection biases, such as over-policing in minority neighborhoods, result in higher bias in urban data \citep{franklin2024sociodemographic}. 
For example, crime reporting disparities often arise not from higher crime rates but from focused police surveillance in certain communities. 
These data collection biases are inherently difficult to mitigate. 
However, models that focus solely on prediction accuracy can amplify these biases, leading to over-prediction of crime events and attracting even more surveillance, thereby exacerbating the disparity \citep{Dong_Liu_Jalaian_Li_2022}.

To mitigate these disparities and achieve equality, previous work has integrated demographic information into the model design. 
The majority of models use demographic information as auxiliary information within the model architecture and loss function designs \citep{zhang2023enhancing} during training ST-GNN models. 
However, in practice, factors such as privacy and regulation often preclude the collection of protected features or their use for training or inference, severely limiting the applicability of traditional fairness research. 
Moreover, models trained with demographic information directly use this data to guide the training, without fully addressing where the disparity originates within the ST-GNN model, such as the message-passing mechanism \citep{zheng2023fairness}.
Therefore, we ask: \textbf{How can we train an ST-GNN based urban prediction model to improve equality when we do not even know the protected group memberships}?

In this paper, we address algorithmic equality and reduce prediction disparities by focusing on model design without incorporating external demographic information. 
We aim to mitigate spatial disparities from ST-GNN models, reduce spatial segregation, and thus address demographic disparities.
Our methodology includes the Residual-Aware Attention (RAA) Block and an equality-enhancing loss function. 
The RAA Block adapts the adjacency matrix during training to dynamically adjust spatial relationships based on residuals, reducing local segregation of errors. 
The equality-enhancing loss function integrates the mean squared error with terms that penalize similar residual patterns in adjacent neighborhoods and incorporate spatial clustering or information redundancy metrics like Moran's I or Generalized Entropy Index (GEI). 
This approach identifies and mitigates the sources of disparity within the model itself, ensuring urban prediction models do not further entrench existing inequities, thus supporting more balanced and just urban planning and policy-making.

We conducted a case study on urban prediction tasks in Chicago using datasets including travel demand, crime, and accident reports. 
Our model achieved a 48\% significant improvement in fairness metrics while only incurring a 9\% increase in error metrics. 
Spatial analysis of residual distributions revealed that models with RAA Blocks produced more equitable prediction results, particularly by reducing errors clustered in central regions. 
Case studies of different community areas demonstrated how the attention mechanism within the RAA Block effectively adjusted focus to address spatial and demographic disparities. 
The attention maps illustrated that our model identifies and emphasizes relationships between regions with similar characteristics, leading to more balanced predictions. 
Overall, our methodology significantly reduces disparities in urban prediction models with minimal impact on accuracy, supporting more equitable urban planning and policy-making.

% \hanyong{\textbf{[Insert a figure illustrating the distribution of prediction residuals and their correlation with income/racial group distributions.]}} 

Our contributions are summarized as follows:
\begin{enumerate}
    \item We introduce residuals as indicators of fairness in predictions, demonstrating their utility in highlighting disparities in ST-GNN model outputs.
    \item We design the RAA Block and propose an equality-enhancing loss function that incorporates spatial and demographic disparity metrics. This function balances prediction accuracy with fairness, ensuring that urban prediction models do not entrench existing spatial disparities. We modify the message-passing mechanism in GNNs to ensure the propagation of information that reduces bias.
    \item We conduct a comprehensive case study applying our method to urban prediction tasks in Chicago using a travel demand dataset as an example. 
    % datasets like travel demand, crime, and accident reports. 
    Our model identifies and mitigates prediction disparities without relying on demographic data, leading to more balanced resource allocation and fairer urban planning decisions.
\end{enumerate}

% ===Section...

\section{Literature Review}
\subsection{Urban Prediction with GNNs}
% \hanyong{Find some relevant literatures related to urban prediction, including urban demand, urban event (e.g. crime) data. Please ensure those models predict with GNN or other Deep learning models. Also, you can talk about the }

In recent years, with the advancement of GNN representing spatial data as graphs and its superiority of capturing both spatial and temporal dependencies, we have seen more research in using GNNs to make predictions in the urban computing context \cite{Jin_Liang_Fang_Shao_Huang_Zhang_Zheng_2023, Ye_Zhao_Ye_Xu_2022}. One major field of application is within transportation. Various studies leverage GNNs to make demand predictions in the context of ride-hailing \cite{Ke_Zheng_Yang_Chen_2017, Wang_Yin_Chen_Wo_Xu_Zheng_2019,  Ye_Sun_Du_Fu_Xiong_2021, Geng2019SpatiotemporalForecasting, Wang_Yin_Chen_Liu_Wang_Wo_Xu_2021, Jin_Xi_Sha_Feng_Huang_2022, Jin_Cui_Zeng_Tang_Feng_Huang_2020}. 
For example, \citet{Wang_Yin_Chen_Wo_Xu_Zheng_2019} solves the Origin-Destination Matrix Problem via introducing grid-embedding and multi-task learning modules to capture both spatial and temporal attributes. \citet{Jin_Cui_Zeng_Tang_Feng_Huang_2020} leverages graph convolutional neural networks in combination with pixel-level representations to capture the joint latent distribution of ride-hailing demands. Other kinds of prediction focus on traffic incident \cite{Yu_Du_Hu_Sun_Han_Lv_2021, Jin_Liu_Li_Huang_2023, Wang_Lin_Guo_Wan_2021}, travel time \cite{Fang_Huang_Wang_Zeng_Liang_Wang_2020, Huang_Huang_Fang_Feng_Chen_Liu_Yuan_Wang_2022}, human mobility \cite{Wang_Jiang_Xue_Salim_Song_Shibasaki_2021}, traffic flow \cite{Yang_Li_Qi_2024,Lu_Gan_Jin_Fu_Wang_Zhang_2022, Tian_Wang_Hu_Ma_2023, wu2021inductive, fang2021spatial}, and trajectory \cite{Peng_Zhang_Li_Zheng_2021, Mohamed_Qian_Elhoseiny_Claudel_2020, Liu_Yao_Li_Wang_Sammut_2022, Sighencea_Stanciu_Căleanu_2023}. 
\citet{Yu_Du_Hu_Sun_Han_Lv_2021} proposed a framework combining spatial graph convolutional network, spatiotemporal standard convolutions, and the embedding layers to capture spatiotemporal and external features for traffic accident prediction. \citep{Wang_Jiang_Xue_Salim_Song_Shibasaki_2021} incorporated multimodal-driven heterogeneous mobility information network and external event-driven memory-augmented dynamic filter generator into the ST-Net to understand crowd movements. \citep{Derrow_She_Wong_Lange_Hester_Perez_Nunkesser_Lee_Guo_Wiltshire_2021} utilized GNNs into real-world ETA estimation in Google Maps, improving the model application with design and operation strategies such as Meta-Gradients and semi-supervised training. 
\citet{fang2021spatial} improved GNNs' depth and capacity in order to extract long-range spatiotemporal dependencies while leveraging both spatial and semantic neighbor nodes in traffic flow prediction.

Apart from prosperous applications in transportation, GNNs are also well developed in many other fields in urban computing such as public safety and environment \cite{Jin_Liang_Fang_Shao_Huang_Zhang_Zheng_2023}. 
GNNs have been popular in crime prediction due to the spatiotemporal correlations of crime incidences \cite{Xia_Huang_Xu_Dai_Bo_Zhang_Chen_2021, Sun_Zhou_Tian_Liao_Xie_2022, Zhang_Cheng_2020, Wang_Lin_Yang_Sun_Yue_Shahabi_2022}. 
Specifically, \citet{Zhang_Cheng_2020} introduced a framework with a gated network and localized diffusion network to explain both temporal and spatial propagations in network-level crime prediction. 
\citet{Wang_Lin_Yang_Sun_Yue_Shahabi_2022} designed a network combining an adaptive region graph learning module, a homophily-aware constraint, and a gated recurrent unit in the diffusion layer that learns correlations based on both distance and crime pattern similarities. 
In remote sensing, \citet{Zhou_Wang_Ding_Liu_Weng_Xiao_2023} proposed a Siamese graph convolutional network to detect urban multi-class change. 
Similarly, GNNs also embraced a wide adoption in the field of atmospheric and disaster predictions  \cite{Liang_Ke_Zhang_Yi_Zheng_2018, Zhou_Zhang_Du_Liu_2021, Farahmand_Xu_Mostafavi_2023}.

Despite the abundance usage of the usage of GNNs in urban computation, existing research only focus on reaching better prediction accuracy levels and ignores the prediction unfairness issues resulting from the algorithm designs. 

\subsection{Algorithmic Fairness}
It is widely accepted that there are two different fairness goals, namely \textit{equality} and \textit{equity} \cite{mehrabi_survey_2022}. 
The former describes each unit of analysis receiving the same amount of resources, while the latter describes each unit of analysis as given enough resources for them to succeed \cite{mehrabi_survey_2022}. Similarly, in transportation, there are \textit{vertical equity} and \textit{horizontal equity} describing the same concepts \cite{litman_evaluating_nodate, yan_fairness-aware_2020}. These binary end goals are evaluated through either Disparate Treatment Analysis, which seeks fair treatment, or Disparate Impact Analysis, which aims for fair impact or results  \cite{mehrabi_survey_2022, pessach_review_2022,caton_fairness_2020}, in algorithmic fairness literature. In this study, we will adopt \textit{equality} or \textit{horizontal equity} as our end goal, aiming for similar treatment for all people and focusing on the equitable sharing of public resources. To evaluate the disparities, we will focus on prediction residuals, the most straightforward measure to reflect under-prediction and over-prediction treatments of the algorithm \citep{kallus2018residual}. 

% rawlsian fairness and the paper
% entropy based improvement on attention mechanism

There are two lines of thought under the umbrella of the Disparate Treatment Analysis: \textit{fairness through awareness} and \textit{fairness through unawareness} \cite{mehrabi_survey_2022}. The former describes achieving similar level results regardless of protected attribute usage during the modeling process, while the latter avoids using such attributes in the decision-making process \cite{Dwork_Hardt_Pitassi_Reingold_Zemel_2012, Kusner_Loftus_Russell_Silva_2017, Grgic-Hlaca_Zafar_Gummadi_Weller}. 
% 2. application: researches that pursue such goals
% 3. demographic un-awareness in prediction
In the field of transportation systems, there have been emerging studies analyzing and mitigating algorithmic disparities. \citet{zheng_equality_2021} highlighted the unfairness in Deep Neural Networks and Discrete Choice Models and proposed a disparity mitigation solution using absolute correlation regularization. Studies by \citet{zheng_fairness-enhancing_2023}, \citet{zhang_enhancing_2023}, and  \citet{Guo_Xu_Zhuang_Zheng_Zhao_2023} addressed demand prediction fairness in ride-hailing and aimed to improve group fairness in prediction results, while the latter also considers downstream operational impacts. 
However, they all sought to achieve \textit{fairness through awareness} by integrating demographic information within the model design, without addressing the origin of disparity within the ST-GNN model, such as the message-passing mechanism. 
Furthermore, privacy and regulation issues often limit the collection and use of protected features for training or inference, reducing the applicability of traditional fairness research. Hence, we explore options to achieve \textit{fairness through unawareness} in our model design.
% Hence, we ask: \textbf{How can we train an ST-GNN-based urban prediction model to improve equality without knowing the protected group's demographic information?}

% Therefore, while algorithmic fairness encompasses all aspects of ensuring equitable outcomes from algorithmic systems, we focus on mitigating prediction disparity without protected attributes in the data. This allow us to specifically target the disproportion of prediction errors across spatial and demographic lines to achieve equitable outputs. To evaluate the disparities, we will focus on prediction residuals, the most straightforward measure to reflect under-prediction and over-prediction treatments of the algorithm \citep{kallus2018residual}. 
% By addressing the spatially segregated prediction residual patterns, we aim to achieve equality in urban prediction models.

\section{Problem Description and Preliminaries}
\label{sec:problem}
\subsection{ST-GNN for Forecasting}

% The critical aspect of ST-GNNs is the introduction of a graph to represent the urban structure, enabling the model to learn spatial dependencies. 
% In our notation, we use calligraphic fonts to indicate sets, bold capital fonts for matrices, subscripts to denote node and time indices, and bold lowercase fonts for vectors. 

We mathematically formulate ST-GNNs in this context as they are extensively utilized in urban prediction tasks \cite{wu2021inductive,zhuang2022uncertainty,Han2019PredictingNetworks,Kong2020STGAT:Forecasting}.
Let the graph be denoted as $\mathcal{G} = (\mathcal{V}, \mathcal{E}, \mathbf{A})$, where $\mathcal{V}$ represents the set of nodes (locations/regions), $\mathcal{E}$ denotes the set of edges, and $\mathbf{A} \in \mathbb{R}^{|\mathcal{V}| \times |\mathcal{V}|}$ is the adjacency matrix that describes the relationships between nodes. 

In this study, we utilize spatiotemporal data collected at the city level, defining regions or census tracts as nodes, and establishing edges based on the geographic distances between regions. Consequently, the adjacency matrix $\mathbf{A}$ reflects the geographical affinities of the regions, with shorter distances indicating larger adjacency values.

We denote the urban spatiotemporal dataset inputs as $\mathcal{X} \in \mathbb{R}^{|\mathcal{V}| \times t}$, where $t$ is the number of time steps. The objective is to predict the target value $\mathcal{Y}_{1:|\mathcal{V}|, t:t+k}$ for the future $k$ time steps, given all past data up to time $t$, denoted as $\mathcal{X}_{1:|\mathcal{V}|, 1:t}$. The goal of the ST-GNN models is to design a model $f_{\theta}$, parametrized by $\theta$, such that:
\begin{equation}
    \hat{\mathcal{Y}}_{1:|\mathcal{V}|, t:t+k} = f_{\theta}(\mathcal{X}_{1:|\mathcal{V}|, 1:t}; \mathcal{G}) = f_{\theta}(\mathcal{X}_{1:|\mathcal{V}|, 1:t}; \mathcal{V}, \mathcal{E}, \mathbf{A}),
    \label{eq:GNN}
\end{equation}
where $\hat{\mathcal{Y}}_{1:|\mathcal{V}|, t:t+k}$ is the predicted target value, typically $\mathcal{X}_{1:|\mathcal{V}|, t:t+k}$ in forecasting tasks. The performance of the forecasting results is measured using residual-based metrics such as Root Mean Squared Error (RMSE) and Mean Absolute Error (MAE). The prediction residuals are defined as $\textbf{r}_{1:|\mathcal{V}|, t:t+k} = \mathcal{Y}_{1:|\mathcal{V}|, t:t+k} - \hat{\mathcal{Y}}_{1:|\mathcal{V}|, t:t+k}$. We denote $\textbf{r}_{1:|\mathcal{V}|, t:t+k}$ as $\mathbf{r}$ for short.
The loss function used to optimize the model during training aims to minimize RMSE or MAE to ensure prediction accuracy.

However, few studies have examined whether these accuracy-oriented approaches are equitable for urban planning. Prediction biases may vary across different spatial regions or demographic groups, potentially leading to inequitable transportation management and urban planning.

\subsection{Measurement of Disparity in ST-GNN Prediction Results}
The most straightforward measurement of disparities in ST-GNN prediction results is the prediction residuals. 
Disparities are defined in terms of spatial and demographic factors.

% \hanyong{Change black and white to minority and majority}

Our models consider spatial disparities in order to address demographic disparities. 
Spatial disparity refers to the prediction residuals exhibiting similar over-prediction or under-prediction patterns within local neighborhoods, resulting in significant differences across the global space. 
For instance, the central of Chicago is consistently over-predicted while the southern part is under-predicted, as shown in Figure \ref{fig:vanilla_residual}, which leads to structural differences that can affect resource allocation. 
Over-predicted areas might receive more mobility resources, while under-predicted areas receive less, leading to equality issues.

Mathematically, multiplying a vector by the adjacency matrix $\mathbf{A}$ aggregates information from neighboring nodes.
Thus, we define the spatial disparity $D_s$ based on the sign-aware residual variance of the neighboring nodes' prediction, as shown in Equation \ref{eq:spatial_disparity}:
% \begin{equation}
% \left\{
% \begin{aligned}
%     &\mathbf{r} = \hat{\mathcal{Y}}_{1:|\mathcal{V}|, t:t+k} - \mathcal{Y}_{1:|\mathcal{V}|, t:t+k}, \\
%     &\mathbf{r}^+_i = \max(\mathbf{r}_i, 0), \quad \mathbf{r}^-_i = \max(-\mathbf{r}_i, 0), \\
%     &\mathbf{s}^+ = \mathbf{A} \mathbf{r}^+, \quad \mathbf{s}^- = \mathbf{A} \mathbf{r}^-, \\
%     &\bar{\mathbf{s}}^+ = \frac{1}{|\mathcal{V}|} \sum_{i=1}^{|\mathcal{V}|} \mathbf{s}^+_i, \quad \bar{\mathbf{s}}^- = \frac{1}{|\mathcal{V}|} \sum_{i=1}^{|\mathcal{V}|} \mathbf{s}^-_i, \\
%     &\sigma^2_{s^+} = \frac{1}{|\mathcal{V}|} \sum_{i=1}^{|\mathcal{V}|} (\mathbf{s}^+_i - \bar{\mathbf{s}}^+)^2, \quad \sigma^2_{s^-} = \frac{1}{|\mathcal{V}|} \sum_{i=1}^{|\mathcal{V}|} (\mathbf{s}^-_i - \bar{\mathbf{s}}^-)^2, \\
%     &D_s = \sigma^2_{s^+} + \sigma^2_{s^-}.
% \end{aligned}
% \right.
% \label{eq:spatial_disparity}
% \end{equation}
\begin{equation}
D_s = \frac{1}{|\mathcal{V}|} \sum_{i=1}^{|\mathcal{V}|} \Big[ (\mathbf{A} \mathbf{r}^+_i - \bar{\mathbf{s}}^+)^2 + (\mathbf{A} \mathbf{r}^-_i - \bar{\mathbf{s}}^-)^2 \Big],
\label{eq:spatial_disparity}
\end{equation}
where $\mathbf{r}^+$ and $\mathbf{r}^-$ represent the positive and negative residuals, $\mathbf{s}^+$ and $\mathbf{s}^-$ are the spatially weighted residuals, and $\bar{\mathbf{s}}^+$ and $\bar{\mathbf{s}}^-$ are their respective averages. The adjacency matrix $\mathbf{A}$ captures the influence of neighboring nodes, and $D_s$ summarizes variance in residuals across neighbors.

% In these equations, $\mathbf{r}$ represents the residuals, $\mathbf{r}^+$ and $\mathbf{r}^-$ represent the positive and negative residuals (over-predictions and under-predictions), respectively. 
% The adjacency matrix $\mathbf{A}$ is used to weight the residuals spatially, capturing the influence of neighboring nodes. 
% $\mathbf{s}^+$ and $\mathbf{s}^-$ are the spatially weighted residuals, and $\sigma^2_{s^+}$ and $\sigma^2_{s^-}$ represent their variances. 
% The spatial disparity $D_s$ is the sum of these variances to show the differences in prediction residuals among neighboring regions.

Another type of disparity arises when over-prediction is positively correlated with specific demographic groups. A positive correlation implies that prediction residuals favor the majority over the minority group. 
This paper focuses on black and white populations as the minority and majority groups, respectively. Disparities with respect to demographic groups can also be reflected in spatial disparities, especially in cases where different demographic groups are spatially segregated, as seen in our case study area: Chicago, see Figure \ref{fig:minority_rate}. We define the demographic disparity as Equation \ref{eq:demo_disparity}:
\begin{equation}
    D_d = \left| \text{Corr}(\mathbf{r}, Pop_{minor}) |  + | \text{Corr}(\mathbf{r}, Pop_{major}) \right|,
\label{eq:demo_disparity}
\end{equation}
where $\text{Corr}(\mathbf{r}, Pop_{minor})$ and $\text{Corr}(\mathbf{r}, Pop_{major})$ represent the Pearson correlation coefficients between the residuals $\mathbf{r}$ and the minority ($Pop_{minor}$) and majority population percentages ($Pop_{major}$), respectively.

\subsection{Attention Mechanism Preliminary}
\label{sec:attention}

The attention mechanism enhances neural networks by focusing on relevant parts of the input, enabling models to capture dependencies in sequential and spatial data effectively.

Given an input data matrix $\mathbf{X} \in \mathbb{R}^{n \times d}$, where $n$ is the number of elements (e.g., time steps, nodes) and $d$ is the feature dimension, the query (\textbf{Q}), key (\textbf{K}), and value (\textbf{V}) matrices are computed as $\mathbf{Q} = \mathbf{X} \mathbf{W}_Q$, $\mathbf{K} = \mathbf{X} \mathbf{W}_K$, and $\mathbf{V} = \mathbf{X} \mathbf{W}_V$, where $\mathbf{W}_Q, \mathbf{W}_K, \mathbf{W}_V \in \mathbb{R}^{d \times d_k}$ are learnable parameters, and $d_k$ is the feature dimension of the query, key, and value vectors.
The attention scores are calculated using the scaled dot-product of the query and key matrices as:
\begin{equation}
    \mathbf{S} = \frac{\mathbf{Q} \mathbf{K}^\top}{\sqrt{d_k}},
    \label{eq:attention}
\end{equation}
where $\mathbf{S} \in \mathbb{R}^{n \times n}$ quantifies the relevance between input elements, with higher scores indicating stronger relationships.

The attention weights are obtained by applying the softmax function to normalize the scores: $\mathbf{H} = \text{softmax}(\mathbf{S})$, where $\mathbf{H}$ represents the importance of each element in the input.

This mechanism has become a cornerstone of modern neural networks, significantly improving performance in tasks requiring an understanding of complex spatial and sequential relationships.

% where $\mathbf{A} \in \mathbb{R}^{n \times n}$ represents the attention weights, ensuring that they sum to one across each row. 
% These weights determine the importance of each element in the input data relative to the others.
% The final output of the attention mechanism is computed as a weighted sum of the value vectors, using the attention weights:
% \begin{equation}
%     \text{Attention}(\mathbf{Q}, \mathbf{K}, \mathbf{V}) = \mathbf{A} \mathbf{V}.
% \end{equation}

The practical significance of the attention mechanism lies in its ability to dynamically focus on the most relevant parts of the input data. This capability is crucial for tasks such as natural language processing, where the context of words can vary significantly, and for image recognition, where different parts of an image may hold varying degrees of importance. By learning to prioritize certain elements, the attention mechanism enhances the model's ability to understand and process complex data structures.

The integration of the attention mechanism in neural networks has led to significant improvements in performance and has become a foundational component in many state-of-the-art models, such as the Transformer architecture \citep{vaswani2017attention}. 
The flexibility and effectiveness of the attention mechanism make it an essential tool for advancing the capabilities of machine learning and deep learning models.

\section{Methodology}
\label{sec:method}

\begin{figure}[htbp]
    \centering
    \includegraphics[width=1\linewidth]{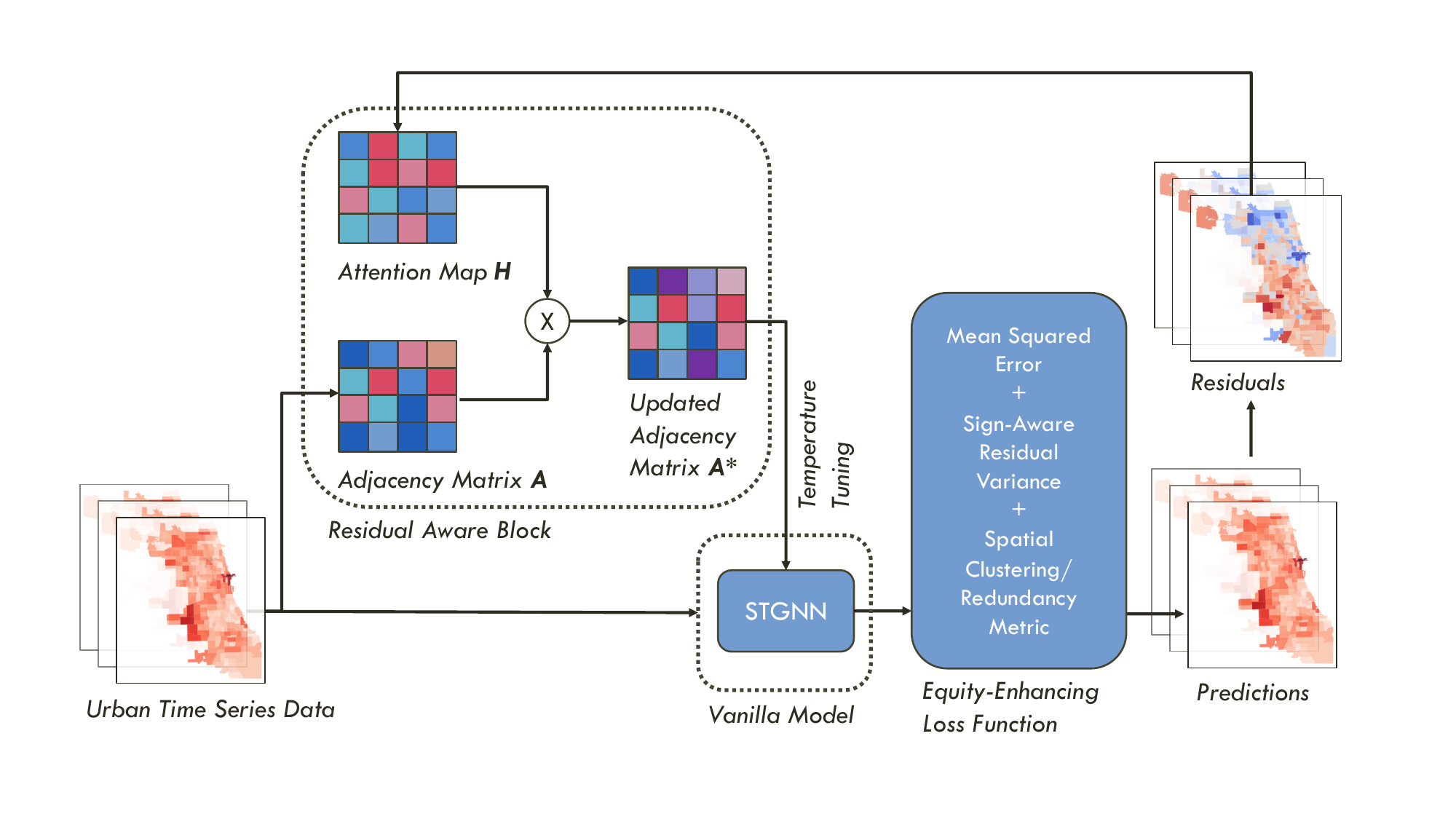}
    \caption{Prediction framework with residual aware attention block.}
    \label{fig:framework}
\end{figure}

Our overall framework, depicted in Figure \ref{fig:framework}, enhances existing ST-GNN architectures for urban prediction by integrating two key components to address disparities: 
1) The RAA Block, which adapts the adjacency matrix during training to reduce spatial disparities in the model outputs. This adaptation leverages the residuals to dynamically adjust the spatial relationships within the graph, focusing less on the local segregation of residuals and errors.
2) An equality-enhancing loss function that penalizes adjacent neighborhoods with similar residual patterns and incorporates a spatial clustering redundancy metric. This function aims to reduce spatial disparities and, consequently, demographic disparities by balancing the importance of prediction accuracy with spatial and demographic equality.

\subsection{RAA Block}
The adjacency matrix in GNN structures is crucial in shaping the GNN results by ensuring that neighboring information is shared, leading to similar feature representations for adjacent nodes. This emphasis on local information can cause spatial disparity, as discussed in Equation \ref{eq:spatial_disparity}.

As introduced in Section \ref{sec:attention}, the attention mechanism is a powerful tool that allows models to focus on specific parts of the data. 
To adaptively mitigate model disparities, we incorporate an RAA Layer using the residuals from each training step as inputs.
We plug in the inputs during training into the equations explained in Section \ref{sec:attention}.

As shown in Figure \ref{fig:framework}, during each training step, we obtain a residual vector $\mathbf{r}$. 
% This vector is then flattened while preserving the batch and sequence dimensions, denoted as $\mathbf{r}_{\text{flat}}$. 
The residuals are used to compute the query (\textbf{Q}), key (\textbf{K}), and value (\textbf{V}) matrices through linear transformations. Activation function $tanh$ is applied to introduce non-linearity and map the values to \([-1, 1]\):
\begin{equation}
    \left\{\begin{aligned}
    \mathbf{Q} &= tanh(\mathbf{W_q}(\mathbf{r})), \\
    \mathbf{K} &= tanh(\mathbf{W_k}(\mathbf{r})), \\
    \mathbf{V} &= tanh(\mathbf{W_v}(\mathbf{r})).
    \end{aligned}\right.
\end{equation}

The attention scores and attention weights are then derived as follows: 
\begin{equation}
    \left\{\begin{aligned}
        \mathbf{S} &= \frac{\mathbf{Q} \mathbf{K}^T}{\sqrt{|\mathbf{K}|}},\\
        \mathbf{H} &= \text{softmax}(\mathbf{S}).
    \end{aligned}\right.
\end{equation}
which indicates the relative importance of each element.

The adapted adjacency matrix is then formed by applying an element-wise Hadamard product between the original adjacency matrix and the attention weights: 
\begin{equation}
    \mathbf{A}_{\text{adapted}} = \mathbf{A} \odot \mathbf{H}.
\end{equation}

% Finally, we multiply the adjusted adjacency matrix element-wise with the original adjacency matrix to form the adapted adjacency matrix: $\mathbf{A} \odot \mathbf{H}$, where $\odot$ denotes the Hadamard product.

This adapted adjacency matrix is used in the next epoch of training, allowing the model to treat the adapted spatial relationships within the graph, thereby focusing less on the segregation of residuals and errors in local communities. 
By integrating this adaptive attention mechanism, our approach dynamically adjusts the model's treatment of spatial relationships, reducing spatial and demographic disparities in urban predictions.

\subsection{Equality-enhancing Loss Function }
To address disparities and achieve equality in urban prediction, we also propose the equality-enhancing loss function. 
The loss function integrates the mean squared error (MSE) loss with additional terms that account for spatial disparities. 
The overall loss function is:
\begin{equation}
    \mathcal{L}_{\text{joint}} = \mathcal{L}_{\text{prediction}} + \lambda_s D_s + \lambda_d D_d,
    \label{eq:loss_overall}
\end{equation}
where $\mathcal{L}_{\text{prediction}}$ is the MSE loss, and $\lambda_s$ and $\lambda_d$ are regularization parameters balancing the importance of the terms, which by default is 0.05 and is tunable according to different datasets.
This approach ensures that urban prediction models do not further entrench existing inequities, thereby supporting more equitable and just urban planning and policy-making.

% the sign-aware residual variance term that aims to reduce the imbalance of over-prediction and under-prediction by considering the variance of spatially weighted residuals:
% \begin{equation}
%     D_s = \mathbb{E}[(\mathbf{A} \cdot \mathbf{r}_{+} - \mathbb{E}(\mathbf{r}_{+}))^2] + \mathbb{E}[(\mathbf{A} \cdot \mathbf{r}_{-} - \mathbb{E}(\mathbf{r}_{-}))^2],
%     \label{eq:Ds}
% \end{equation}
% where $\mathbf{r}_{+}$ and $\mathbf{r}_{-}$ are positive and negative residuals, respectively.

Specifically, the $D_s$ is the sign-aware residual variance term that measures spatial disparity with the same definition as Equation \ref{eq:spatial_disparity}.
The $D_d$ term utilizes fairness metrics like Moran's I or GEI to measure spatial clustering or information redundancy, focusing on reducing the overall spatial unevenness of the prediction residuals. The detailed formulation of these metrics is outlined in Section \ref{sec:fairness_metrics}.

\section{Experiments}
\label{sec:experiment}
\subsection{Urban Data Collections in Chicago}

The demographic dataset for this research is sourced from the American Community Survey (ACS) for the years 2017-2018. The ACS provides detailed sociodemographic information, including income brackets, age demographics, racial compositions, commuting modes, and average commuting durations. This data is specific to each of the 811 census tracts within Chicago \citep{zhuang2024advancing,wang2024deep}. 
Key attributes include total population, age groups, racial composition, education levels, economic status, and travel information such as commuting times\footnote{\url{https://www.census.gov/programs-surveys/acs/data.html}}.

We use \textbf{Chicago Data Portal (CDP)} \footnote{\url{https://data.cityofchicago.org/Transportation/Transportation-Network-Providers-Trips/m6dm-c72p}} to evaluate the effects of our model. CDP contains the trip records of Transportation Network Providers (ride-sharing companies) in the Chicago area. The city of Chicago is divided into 77 zones and the trip requests with pick-up and drop-off zones are recorded every 15 minutes. We use 4-month observations from September 1st, 2019 to December 30th, 2019. 

% We use three urban datasets in Chicago:

% (1) \textbf{Chicago Traffic Crash Data (CTC)} \footnote{\url{https://data.cityofchicago.org/Transportation/Traffic-Crashes-Crashes/85ca-t3if}} sourced from 277 police beats between January 1, 2016 and January 1, 2023. The CTC data records show information about each traffic crash on city streets within the City of Chicago limits and under the jurisdiction of the Chicago Police Department. 

% (2) \textbf{Chicago Crime Records (CCR)} \footnote{\url{https://data.cityofchicago.org/Public-Safety/Crimes-2001-to-Present/ijzp-q8t2}} derived from 77 census tracts spanning January 1, 2003 to January 1, 2023. This dataset reflects reported incidents of crime (with the exception of murders where data exists for each victim) that occurred in the City of Chicago.

% (3) \textbf{Chicago Data Portal (CDP)} \footnote{\url{https://data.cityofchicago.org/Transportation/Transportation-Network-Providers-Trips/m6dm-c72p}} contains the trip records of Transportation Network Providers (ride-sharing companies) in the Chicago area. The city of Chicago is divided into 77 zones and the trip requests with pick-up and drop-off zone are recorded every 15min. We use 4-month observations from September 1st, 2019 to December 30th, 2019. 

% \hanyong{Please introduce Chicago and its spatial segregation of demographics. [Insert the figure of the north and south separation]}

We have chosen Chicago as a case study because of its heavy spatial segregation of different racial groups, for which unfair algorithmic prediction results may lead to even worse segregation. 
Fig \ref{fig:minority_rate} shows the racial minority rate in Chicago in 2019. 
It is clear that minority races cluster around the south and middle-west of the city, while regions in the north have far fewer minorities than the other regions.

% \begin{figure}[htbp]
%     \centering
%     \includegraphics[width=0.5\linewidth]{fig/minority_Chicago.pdf}
%     \caption{Distribution of the racial minority groups in Chicago, defined as the percentage of the non-white population in each community area. \redfont{Better say the disadvantaged group?}}
%     \label{fig:minority_rate}
% \end{figure}

\subsection{Model Comparison}
\label{sec:model_comparison}
% \hanyong{Please find the description of the models DCRNN, DGCRN, STGCN, and so on}

We selected four prevalent ST-GNN models as base models to demonstrate the effectiveness of our proposed method:

\begin{itemize}
    \item \textbf{Diffusion Convolutional Recurrent Neural Network (DCRNN)} by \citet{Li2018DiffusionForecasting} models traffic dynamics using diffusion convolution. It captures spatial dependencies with bidirectional random walks on the graph and temporal dependencies with an encoder-decoder architecture and scheduled sampling.
    
    \item \textbf{Dynamic Spatial-Temporal Aware Graph Neural Network (DSTAGNN)} by \citet{lan2022dstagnn} learns dynamic association attributes from data to represent the graph. It uses a multi-head attention mechanism for spatial variances and handles temporal dependencies with features of multi-receptive fields.
    
    \item \textbf{Spatio-Temporal Graph Convolutional Network (STGCN)} by \citet{Yu2018Spatio-TemporalForecasting} features two spatio-temporal convolutional blocks that leverage graph convolutional layers to capture spatial dependencies and temporal gated convolution layers for temporal dynamics.
    
    \item \textbf{Adaptive Graph Convolutional Recurrent Network (AGCRN)} by \citet{bai2020adaptive} consists of two modules: Node Adaptive Parameter Learning, which learns node-specific parameters from node embeddings, and Data Adaptive Graph Generation, which generates a graph from the training data. This architecture captures fine-grained variability in space and time.
\end{itemize}

For each model, the performance of the vanilla version will be compared with the performance of the models with the RAA block and enhanced loss functions added. 
The vanilla version model implementations we used were based on the repository by \citet{liu2023largest}\footnote{\url{https://github.com/liuxu77/LargeST/}}.

% \hanyong{Introduce what different ablation module you have made, for example, the }

\subsection{Fairness Metrics}
\label{sec:fairness_metrics}
% \hanyong{What are the fairnessmetrics we use to compare. like GEI, moran's I}

Fairness metrics are vital for evaluating equality in urban prediction models, addressing both spatial and demographic disparities. 
These metrics collectively help us ensure that our models do not exacerbate existing disparities, promoting fair and equitable urban predictions. We selected GEI, SDI, and Moran's I to evaluate three aspects of the residual distribution: entropy, correlation to the demographics features, and spatial clusteringness.

\subsubsection{Generalized Entropy Index (GEI)}
GEI is a metric developed by \citep{Speicher_Heidari_2018} used to compare unfair algorithmic treatments to individuals in a population, originating from existing economic inequality indices. 
It reflects spatial disparity rather than demographic disparity.
% In this research, we calculate the GEI of prediction residuals. A smaller value represents a more even distribution between all residuals at each unit of analysis and the residual mean, hence fairer prediction. The exact formulation is as follows:
% \begin{align}
%     & b_i = r_i + m \text{, where } m \in \mathbb{R}^+ s.t. \forall \textbf{r}, b_{i} \geq 0,  \\
%     & GEI = \frac{1}{|\mathcal{V}|\alpha(\alpha-1)}\sum_{i=1}^{|\mathcal{V}|} [(\frac{b_i}{\Bar{b}})^\alpha - 1],
% \end{align}
% where $r_i^t$ means the residual at node $i$. We first convert all residuals to non-negative numbers by adding a constant. We also use a constant $\alpha$ that controls how much weight to put on larger residuals and set it as $\alpha = 2$.

We calculate the GEI of prediction residuals, where a smaller value indicates a more even distribution of residuals across all units of analysis, resulting in fairer predictions. Denote a non-negative transformation of the residuals $b_i = r_i + m, \text{where } m \in \mathbb{R}^+ \text{ such that } b_i \geq 0 \; \forall i$, and $r_i$ represents the residual at node $i$. The formulation is as follows:
\begin{equation}
    GEI = \frac{1}{|\mathcal{V}|\alpha(\alpha-1)}\sum_{i=1}^{|\mathcal{V}|} \left[\left(\frac{b_i}{\Bar{b}}\right)^\alpha - 1\right],
\end{equation}
The parameter $\alpha$ controls the emphasis on larger residuals, and we set $\alpha = 2$ in this research.

\subsubsection{Moran's I}
Moran's I is a metric in spatial statistics used to measure spatial auto-correlations. 
It measures spatial disparity from the overall clustering of the data spatial distribution, ranging from -1 to 1. 
A value of 1 means total positive spatial auto-correlation, meaning similar values are spatially clustered together. 
A value of 0 means random distribution. 
A value of -1 means total negative spatial auto-correlation, meaning similar values are dispersed. 
Since we hope to avoid spatial clustering, a lower value of Moran's I is appreciated more than higher values.
% The Moran's I is formulated as follows:
% \begin{equation}
%     \left\{\begin{split}
%     & W = \sum_{i=1}^N \sum_{j=1}^N a_{ij}, \\
%     & I = \frac{N}{W} \frac{\sum_{i=1}^{N} \sum_{j=1}^{N} a_{ij} (r_i - \overline{r}) (r_j - \overline{r})}{\sum_{i=1}^{N} (r_i - \overline{r})^2},\\
%     & I^* = I + 1,
%     \end{split} \right.
% \end{equation}
% where $r_i$ and $r_j$ means the residual at node $i$, $a_{ij}$ are the elements of the adjacency matrix $\mathbf{A}$, and $N$ is the number of row or columns of the adjacency matrix $\mathbf{A}$. We added 1 to the final Moran's I to avoid a negative value in the loss function.

The weight $W$ is computed as:
$ W = \sum_{i=1}^N \sum_{j=1}^N a_{ij}, $
where $a_{ij}$ are elements of the adjacency matrix $\mathbf{A}$, and $N$ is the number of nodes.
Moran's I is then calculated as:
\begin{equation}
    I = \frac{N}{W} \frac{\sum_{i=1}^N \sum_{j=1}^N a_{ij} (r_i - \overline{r}) (r_j - \overline{r})}{\sum_{i=1}^N (r_i - \overline{r})^2},
\end{equation}
where $r_i$ and $r_j$ are residuals at nodes $i$ and $j$, and $\overline{r}$ is the residual mean.
In order to ensure non-negative values for the loss function, we adjust as:
$ I^* = I + 1. $
This adjustment supports penalizing spatial clustering in our loss function.

\subsubsection{Scaled Disparity Index}
% Hence, we propose using the Pearson correlation coefficients between prediction residuals and the minority populations ($\text{Corr}(\mathbf{r}, Pop_{minor})$) and residuals and the majority populations ($\text{Corr}(\mathbf{r}, Pop_{major})$) to assess fairness.

% To quantify disparity, we introduce the Scaled Disparity Index (SDI):
% \begin{equation}
%     SDI = \frac{|\text{Corr}(\mathbf{r}, Pop_{minor}) - \text{Corr}(\mathbf{r}, Pop_{major})|}{|\text{Corr}(\mathbf{r}, Pop_{minor})| + |\text{Corr}(\mathbf{r}, Pop_{major})|} \cdot \sqrt{|\text{Corr}(\mathbf{r}, Pop_{minor}) \cdot \text{Corr}(\mathbf{r}, Pop_{major})|}
% \end{equation}

% The first term of the SDI normalizes the disparity by taking the difference between $\text{Corr}(\mathbf{r}, Pop_{minor})$ and $\text{Corr}(\mathbf{r}, Pop_{major})$ and scaling it by the sum of their absolute values. The geometric mean of $\text{Corr}(\mathbf{r}, Pop_{minor})$ and $\text{Corr}(\mathbf{r}, Pop_{major})$ then adjusts this normalized disparity by accounting for the magnitude of the correlations. This scaling is crucial, as it distinguishes between scenarios where correlations have different magnitudes but similar patterns of disparity.

While we do not incorporate demographic disparities in model training, we still need to measure if mitigating spatial disparities also reduces demographic disparities.
Hence, we propose using the Pearson correlation coefficients between prediction residuals and the minority populations ($\text{Corr}(\mathbf{r}, Pop_{minor})$) and residuals and the majority populations ($\text{Corr}(\mathbf{r}, Pop_{major})$) to assess fairness.

To quantify disparity, we introduce the SDI:
\begin{equation}
    SDI = 
    \frac{|\Delta \text{Corr}|}{|\text{Corr}_{\text{minor}}| + |\text{Corr}_{\text{major}}|} 
    \cdot \sqrt{|\text{Corr}_{\text{minor}} \cdot \text{Corr}_{\text{major}}|},
    \label{eq:sdi}
\end{equation}

The first term of the SDI normalizes the disparity by taking the difference between $\text{Corr}(\mathbf{r}, Pop_{minor})$ and $\text{Corr}(\mathbf{r}, Pop_{major})$ and scaling it by the sum of their absolute values. The geometric mean of $\text{Corr}(\mathbf{r}, Pop_{minor})$ and $\text{Corr}(\mathbf{r}, Pop_{major})$ then adjusts this normalized disparity by accounting for the magnitude of the correlations. This scaling is crucial, as it distinguishes between scenarios where correlations have different magnitudes but similar patterns of disparity.

Ultimately, the SDI yields non-negative values, with smaller values indicating greater fairness in our context. The SDI can also be extended to compare other advantaged and disadvantaged groups, such as evaluating equality between poor and rich communities.

\subsection{Error Metrics}

We also employ common error metrics to evaluate the accuracy of our predictions: the Symmetric Mean Absolute Percentage Error (SMAPE) and the Mean Absolute Error (MAE).

The SMAPE is defined as:
\begin{equation}
\text{SMAPE} = \frac{1}{N} \sum_{i=1}^{N} \frac{2 \left| \hat{y}_i - y_i \right| + \epsilon}{\left| \hat{y}_i \right| + \left| y_i \right| + \epsilon},
\end{equation}
where $\hat{y}_i$ represents the predicted values, $y_i$ represents the true values, $N$ is the number of observations, and $\epsilon$ is a small constant to avoid division by zero. 
SMAPE provides a normalized measure of the average absolute error, considering the scale of the true values.

The MAE is defined as:
\begin{equation}
\text{MAE} = \frac{1}{N} \sum_{i=1}^{N} \left| \hat{y}_i - y_i \right|,
\end{equation}
where $\hat{y}_i$ represents the predicted values and $y_i$ represents the true values. 
MAE measures the average magnitude of the errors in a set of predictions, without considering their direction, providing a straightforward interpretation of prediction accuracy.

For all metrics mentioned (GEI, Moran's I, SDI, SMAPE, MAE), lower values indicate better performance.

\subsection{Experiment setup}
\label{sec:experiment_setup}
% \hanyong{How do you set up the ablation experiment. Use or do not use RAA block. Test on the different loss functions and why do we need to use it}

The experiment explores two parts, the first part compares the accuracy and fairness performance of the original vanilla model and the different variants of RAA-enhanced models.
The second part of the experiment is an ablation study investigating the effects of the RAA block, the sign-aware residual variance, and the Moran's I or GEI metrics on the performance of the model. 

All our experiments are implemented on a machine with Ubuntu 22.04, with Intel(R) Core(TM) i9-10980XE CPU @ 3.00GHz CPU, 128GB RAM, and NVIDIA GeForce RTX 4080 GPU.

\subsection{Results}
\subsubsection{Overall Performance}
% \hanyong{[A large table comparing the different models/model cases, like train with normal loss, gei loss and so on. More importantly, you need to summarize and interpret what you see from the table.]}

We compare three variants in total for all four base models in Section \ref{sec:model_comparison}, all of which include the RAA block in the model architecture. 
The variants differ based on the choice of the $D_d$ term in the loss function. 
They include:% For each model variant, we utilize a random seed. 

\begin{itemize}
    \item adding the RAA block and $D_s$ in the loss function;
    \item additionally, adding the Moran's I metric as the $D_d$ term in the loss function;
    \item alternatively, adding the GEI metric as the $D_d$ term in the loss function.
\end{itemize}

\begin{table}[htbp]
\centering
\scriptsize
% \footnotesize
\setlength{\tabcolsep}{3pt}
\begin{tabular}{l|ccccc|lccccc}
\toprule
{\multirow{2}{*}{Models}} & \multicolumn{5}{c}{\textbf{Original model}} & \multicolumn{6}{c}{\textbf{Residual-Aware Attention Variants}} \\
    \cmidrule(lr){2-6} \cmidrule(lr){7-12}
 & MAE & SMAPE & GEI & SDI & Moran's I   &  Variants   & MAE & SMAPE & GEI & SDI & Moran's I \\
\midrule
\textbf{DCRNN} & 8.092 & 0.458 & 1.28 & 0.2 & 0.182
& RAA block + $D_s$ &
$\cellcolor{green!25}7.492_{\downarrow7\%}$ &
$\cellcolor{red!25}0.478_{\uparrow4\%}$ &
$\cellcolor{green!25}1.106_{\downarrow13\%}$ &
$\cellcolor{green!25}0.16_{\downarrow20\%}$ &
$\cellcolor{green!25}0.068_{\downarrow62\%}$ \\
& & & & & & + loss with Moran's I &
$\cellcolor{red!25}8.911_{\uparrow10\%}$ &
$\cellcolor{red!25}0.555_{\uparrow21\%}$ &
$\cellcolor{green!25}1.148_{\downarrow10\%}$ &
$\cellcolor{green!25}0.183_{\downarrow8\%}$ &
$\cellcolor{green!25}-0.135_{\downarrow174\%}$ \\
& & & & & & + loss with GEI &
$\cellcolor{green!25}7.608_{\downarrow5\%}$ &
$\cellcolor{red!25}0.483_{\uparrow5\%}$ &
$\cellcolor{green!25}1.2_{\downarrow6\%}$ &
$\cellcolor{green!25}0.072_{\downarrow64\%}$ &
$\cellcolor{green!25}0.046_{\downarrow74\%}$ \\

\midrule

\textbf{DSTAGNN} & 8.564 & 0.425 & 1.383 & 0.308 & 0.542
& RAA block + $D_s$ &
$\cellcolor{green!25}8.185_{\downarrow4\%}$ &
$\cellcolor{red!25}0.456_{\uparrow7\%}$ &
$\cellcolor{green!25}1.2_{\downarrow13\%}$ &
$\cellcolor{green!25}0.269_{\downarrow12\%}$ &
$\cellcolor{green!25}0.436_{\downarrow19\%}$ \\
& & & & & & + loss with Moran's I &
$\cellcolor{green!25}8.509_{\downarrow0\%}$ &
$\cellcolor{red!25}0.448_{\uparrow5\%}$ &
$\cellcolor{red!25}2.862_{\uparrow106\%}$ &
$\cellcolor{green!25}0.281_{\downarrow8\%}$ &
$\cellcolor{red!25}0.558_{\uparrow2\%}$ \\
& & & & & & + loss with GEI &
$\cellcolor{red!25}9.555_{\uparrow11\%}$ &
$\cellcolor{red!25}0.447_{\uparrow5\%}$ &
$\cellcolor{green!25}0.434_{\downarrow68\%}$ &
$\cellcolor{green!25}0.238_{\downarrow22\%}$ &
$\cellcolor{green!25}-0.021_{\downarrow103\%}$ \\

\midrule

\textbf{STGCN} & 6.948 & 0.428 & 1.506 & 0.337 & 0.394
& RAA block + $D_s$ &
$\cellcolor{red!25}7.329_{\uparrow5\%}$ &
$\cellcolor{red!25}0.538_{\uparrow25\%}$ &
$\cellcolor{green!25}1.075_{\downarrow28\%}$ &
$\cellcolor{green!25}0.235_{\downarrow30\%}$ &
$\cellcolor{green!25}0.01_{\downarrow97\%}$ \\
& & & & & & + loss with Moran's I &
$\cellcolor{red!25}7.885_{\uparrow13\%}$ &
$\cellcolor{red!25}0.506_{\uparrow18\%}$ &
$\cellcolor{red!25}1.692_{\uparrow12\%}$ &
$\cellcolor{green!25}0.106_{\downarrow68\%}$ &
$\cellcolor{green!25}0.201_{\downarrow48\%}$ \\
& & & & & & + loss with GEI &
$\cellcolor{red!25}9.349_{\uparrow34\%}$ &
$\cellcolor{red!25}0.476_{\uparrow11\%}$ &
$\cellcolor{green!25}0.35_{\downarrow76\%}$ &
$\cellcolor{green!25}0.056_{\downarrow83\%}$ &
$\cellcolor{green!25}-0.108_{\downarrow127\%}$ \\

\midrule

\textbf{AGCRN} & 6.988 & 0.425 & 1.265 & 0.316 & 0.064
& RAA block only &
$\cellcolor{green!25}7.039_{\downarrow0\%}$ &
$\cellcolor{red!25}0.465_{\uparrow9\%}$ &
$\cellcolor{green!25}0.975_{\downarrow22\%}$ &
$\cellcolor{green!25}0.007_{\downarrow97\%}$ &
$\cellcolor{green!25}0.019_{\downarrow70\%}$ \\
& & & & & & + loss with Moran's I &
$\cellcolor{red!25}7.393_{\uparrow5\%}$ &
$\cellcolor{red!25}0.49_{\uparrow15\%}$ &
$\cellcolor{green!25}0.978_{\downarrow22\%}$ &
$\cellcolor{green!25}0.07_{\downarrow77\%}$ &
$\cellcolor{green!25}-0.034_{\downarrow153\%}$ \\
& & & & & & + loss with GEI &
$\cellcolor{red!25}8.419_{\uparrow20\%}$ &
$\cellcolor{red!25}0.49_{\uparrow15\%}$ &
$\cellcolor{green!25}0.373_{\downarrow70\%}$ &
$\cellcolor{green!25}0.084_{\downarrow73\%}$ &
$\cellcolor{green!25}0.04_{\downarrow37\%}$ \\

\bottomrule
\end{tabular}
\caption{Comparison of Models with different Residual-Aware Attention versions}
\label{tab:overall_result}
\end{table}

We have calculated the accuracy and fairness metrics for each model and their variants in Table \ref{tab:overall_result}. 
The base model performance metrics and the RAA variants are shown on the left and right sides of the table. 
The green color highlights metric improvements whereas the red color highlights the opposite. 
The percentages followed by each metric result demonstrate the amount of improvement or declination compared to the base model. 

% 1. fairness metrics improve
% 1.1 which variant has better performance or model dependent?
In general, the introduction of the RAA blocks and the additional equality-enhancing loss functions improve the fairness metrics compared to the base model. 
On average, the GEI, SDI, and Moran's I decreased by 18\%, 47\%, and 80\%, respectively. Among all 12 RAA experiments, 10 of them have reduced metrics for each fairness metric, and all experiments have reduced SDI. This signifies the effectiveness of the RAA modules.
If we take the average of the percentage change for each RAA variant, we get a reduction of 40\%, 37\%, and 67\%, respectively, for each of the three variants. Thus, the variant with the RAA block and the GEI loss function has the best performance for reducing fairness metrics.
If we compare all four GNN models, the DCRNN and the AGCRN models with RAA blocks have the most consistent performance. 
Their GEI, SDI, and Moran's I all improved for all three model variations. 
The AGCRN model's overall fairness metric decrease percentages are also the highest compared to other models.
These improvements evidently showcase that reducing spatial disparities in model design can ultimately reduce demographic disparities.

% 2. trade-off between accuracy and fairness, relative small decrease in accuracy compared to the improved fairness percentages
% 2.1 error metrics improve, rashomon effect
In addition, we observe a trend of accuracy-fairness trade-off. 
Among the 12 experiments, all of them have at least one fairness metric improving while having error metrics decreasing. 
This makes sense since the objective of this project is to reduce prediction error variances while sacrificing the accuracy performance. 
It is worth noting that even though this is the case, our contribution allows much more fairness improvement amount compared to the accuracy loss in terms of the percentage change. The average percentage increase for MAE and SMAPE are 7\% and 12\%, which are much lower than the fairness percentage decrease mentioned above.
It is possible to not sacrifice too much accuracy while greatly improving the prediction equality among all regions. 
Furthermore, there are several cases when adding the RAA blocks and the additional regularization terms in the loss function improves both accuracy and fairness. 
This phenomenon was possible due to the \textit{Model Multiplicity} or \textit{Under-Specification}, meaning that the complex nature of the loss function allows multiple objectives to be met during the training process \cite{D’Amour_Heller_Moldovan_Adlam_Alipanahi_Beutel_Chen_Deaton_Eisenstein_Hoffman_2022, Black_Raghavan_Barocas_2022}.

\subsubsection{Residual Spatial Distribution}

\begin{figure}[htbp]
\centering
\begin{subfigure}{\textwidth}
  \centering
  \includegraphics[width=.95\linewidth]{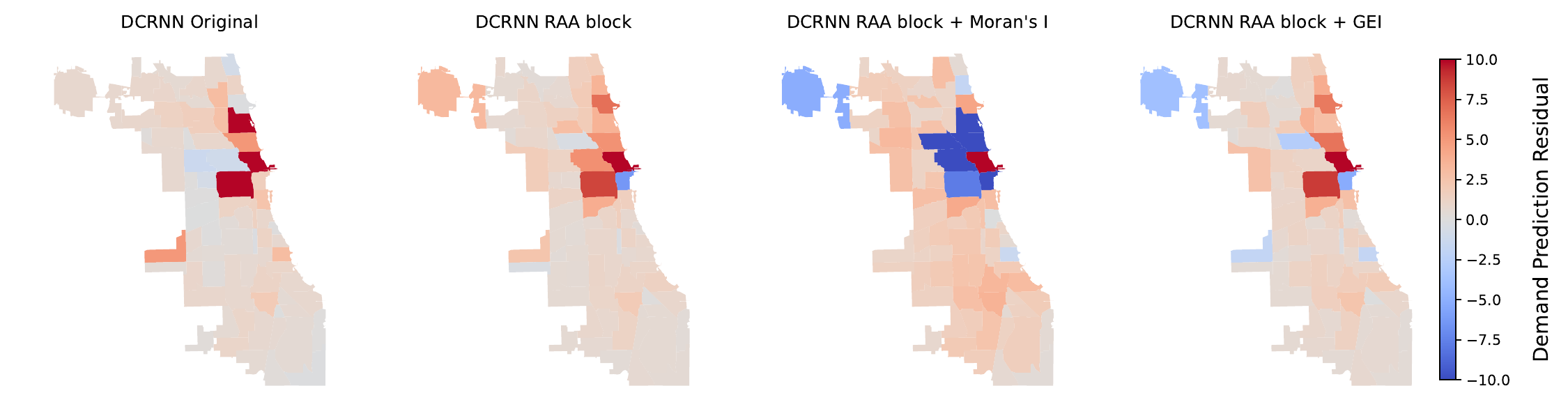}  
  \caption{DCRNN Residual Distribution}
  \label{fig:residual_dcrnn}
\end{subfigure}
\begin{subfigure}{\textwidth}
  \centering
  \includegraphics[width=.99\linewidth]{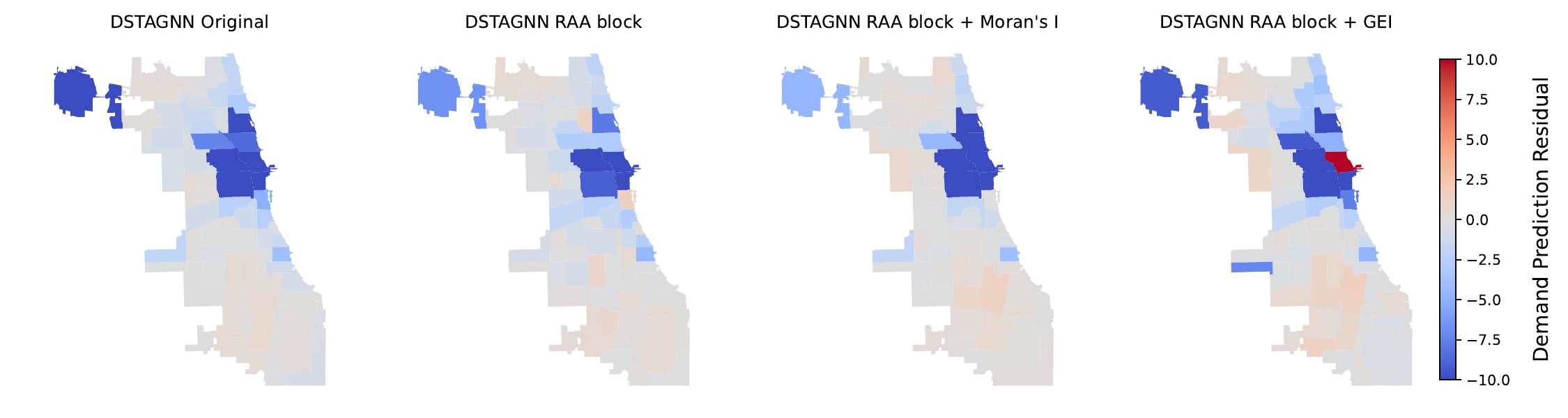}  
  \caption{DSTAGNN Residual Distribution}
  \label{fig:residual_dstagnn}
\end{subfigure}
\begin{subfigure}{\textwidth}
  \centering
  \includegraphics[width=.95\linewidth]{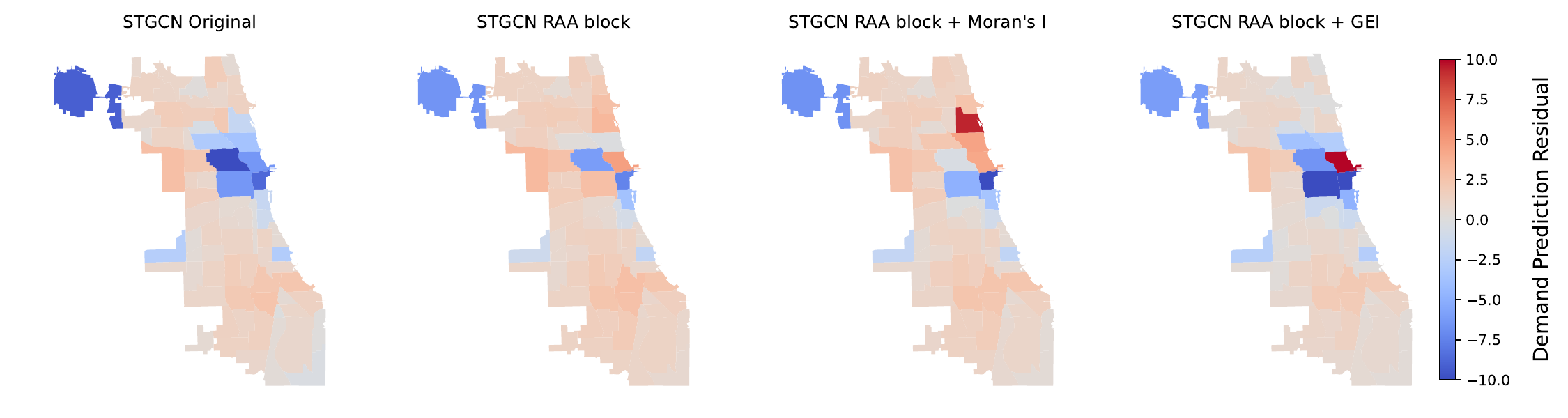}  
  \caption{STGCN Residual Distribution}
  \label{fig:residual_stgcn}
\end{subfigure}
\begin{subfigure}{\textwidth}
  \centering
  \includegraphics[width=.95\linewidth]{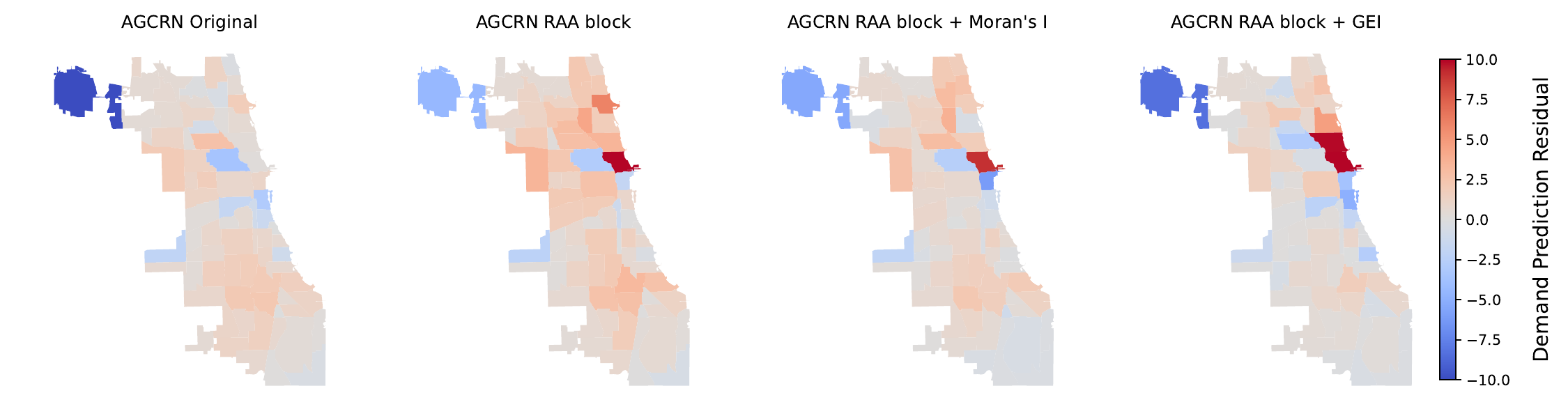}  
  \caption{AGCRN Residual Distribution}
  \label{fig:residual_agcrn}
\end{subfigure}
\caption{Residual spatial distribution in Chicago.}
\label{fig:error_spatial}
\end{figure}

To understand the effect of the RAA modules on the residual spatial distributions, we have plotted the average residuals at each community area on the map, as shown in Figure \ref{fig:error_spatial}. 
The red colors represent positive residuals, or under-prediction, while the blue colors represent the opposite. 
Darker colors demonstrate larger residuals.

% 1 adding the RAA modules are effective
Comparing the spatial distribution of residuals between the original model and models with RAA modules, we observe that the latter produces more equitable prediction results, as indicated by the less prominent residual clusters on the map. 
Models with RAA blocks are effective in reducing errors concentrated in central city regions. For instance, in the case of the STGCN model shown in Figure \ref{fig:residual_stgcn}, significant over-prediction occurs in central Chicago. 
However, models with the RAA block and Moran's I in the loss function display a much smoother residual distribution. 
Although the STGCN model with GEI as a regularization term $D_d$ still exhibits some clustering of large residuals in the central region, the residual signs are more heterogeneous, indicating reduced spatial autocorrelation.

Furthermore, RAA blocks improve residual distribution throughout the city. 
For example, in the AGCRN models shown in Figure \ref{fig:residual_dstagnn}, the original model displays slight under-prediction in the southern part and northern tip of the city. 
This issue is mitigated in models with RAA blocks and additional GEI regularization, resulting in a more equitable distribution of residuals across different regions.

% 2 model specific results, not always work, but in most cases work,
% follow similar trend as the general metric results
It is also worth noting that, although RAA blocks demonstrate some effectiveness in reducing disparities in the residual spatial distribution, their effectiveness is model-dependent. 
There is a trade-off between residual clustering, variance, and the imbalance between over- and under-predictions. 
It is challenging for a single model to improve all three aspects simultaneously, which explains the varied performances of different models on the map and in terms of different fairness metrics.

\subsubsection{Abaltion Study}
An ablation study is a method used in machine learning to evaluate the impact of individual components on a model's performance. 
By systematically altering specific elements and observing changes in performance metrics, researchers can identify which components are essential. 
We conducted 6 ablation cases in total, corresponding to:
\begin{itemize}
    \item \textbf{RAA block}: adding RAA block to the architecture alone;
    \item \textbf{RAA block + $D_s$}: adding the RAA block to the architecture and adding the $D_s$ regularization term in the loss function;
    \item \textbf{RAA block + Moran's I}: adding the RAA block to the architecture and adding the Moran's I metric as the $D_d$ term in the loss function;
    \item \textbf{RAA block + GEI}: adding the RAA block to the architecture and adding the GEI metric as the $D_d$ term in the loss function;
    \item \textbf{RAA block + $D_s$ + Moran's I}: adding the RAA block to the architecture, adding the $D_s$ term and the Moran's I metric as the $D_d$ term to form the Equation \ref{eq:loss_overall};
    \item \textbf{RAA block+ $D_s$ + GEI }: adding the RAA block to the architecture, adding the $D_s$ term and the GEI metric as the $D_d$ term to form the Equation \ref{eq:loss_overall}.
\end{itemize}

% \hanyong{In this study, the ablation study assesses ...... This approach validates design choices, confirming their positive influence on reducing prediction disparities and ensuring the robustness of the model.}

% ===
% \hanyong{Update the moran's I results}

\begin{table}[htbp]
\centering
\begin{tabular}{l|l|ccccc}
\hline
\textbf{No.} & \textbf{Model Variants} & \textbf{MAE} & \textbf{SMAPE} & \textbf{GEI} & \textbf{SDI} & \textbf{Moran's I} \\
\hline
1& Original & 6.948 & 0.428 & 1.506 & 0.337 & 0.394 \\
2& RAA block  & 7.830 & 0.475 & 0.402 & 0.154 & -0.000 \\
3& RAA block + $D_s$ & 7.257 & 0.474 & 1.854 & 0.140 & 0.212 \\
4& RAA block + Moran's I & 10.465 & 0.470 & 0.842 & 0.151 & 0.251 \\
5& RAA block + GEI & 8.028 & 0.525 & 0.525 & 0.044 & -0.136 \\
6& RAA block + $D_s$ + Moran's I & 7.924 & 0.508 & 1.461 & 0.015 & 0.012 \\
7& RAA block + $D_s$ + GEI & 9.349 & 0.476 & 0.350 & 0.056 & -0.108 \\
\hline
\end{tabular}
\caption{Ablation study of our designed modules using the STGCN model.}
\label{table:ablation}
\end{table}

To illustrate the effect of model architecture design choices, we conducted an ablation study in the context of the STGCN model, as shown in Table \ref{table:ablation}. 
The detailed steps are described in section \ref{sec:experiment_setup}. The ablation study validates each strategy we have introduced.

First, the RAA block significantly improves fairness metrics compared to the original model, although it sacrifices accuracy. 
Comparing Model 1 and Model 2 from Table \ref{table:ablation}, we observe reductions in GEI, SDI, and Moran's I, indicating reduced residual variances, correlations to demographics, and spatial autocorrelation. 
This supports the hypothesis that using residuals to calculate the Attention map helps direct the model toward more equitable results.
Moreover, both spatial and demographic disparities could be mitigated using this method.

Second, adding the $D_s$ term improves accuracy but trades off with fairness results. 
This is demonstrated by comparing three pairs of models: Model 2 and Model 3, Model 4 and Model 6, and Model 5 and Model 7. 
Models with $D_s$ regularization show at least one lower accuracy metric compared to those without. 
The $D_s$ term impacts error metrics because it separately calculates the variance of positive and negative residuals, effectively weighting errors more heavily in the loss function, causing the model to focus more on accuracy.

Lastly, adding $D_s$ and $D_d$ terms improves the corresponding fairness metrics. 
Comparing between Model 2 and Model 5, as well as Model 3 and Model 7, we see that introducing GEI in the loss function as a regularization term significantly enhances the GEI performance of model predictions. 
On the other hand, comparing between Model 2 and Model 4 as well as Model 3 and Model 6, we see that the Moran's I metric as the $D_d$ term is less effective in improving the output Moran's I metric, but is able to help with other fairness metrics.
Thus, over-emphasizing any one metric can lead to worse results in others. 
It is crucial to balance multiple objectives in the loss function to avoid such scenarios. 
In our case, this means ensuring similar regularization weights for both accuracy and fairness terms.

\subsubsection{Spatial Attention Distribution}
% \hanyong{[Insert the attention map and the most important areas learned from the attention maps]}

\begin{figure}[htbp]
\centering
\begin{subfigure}{\textwidth}
  \centering
  \includegraphics[width=.99\linewidth]{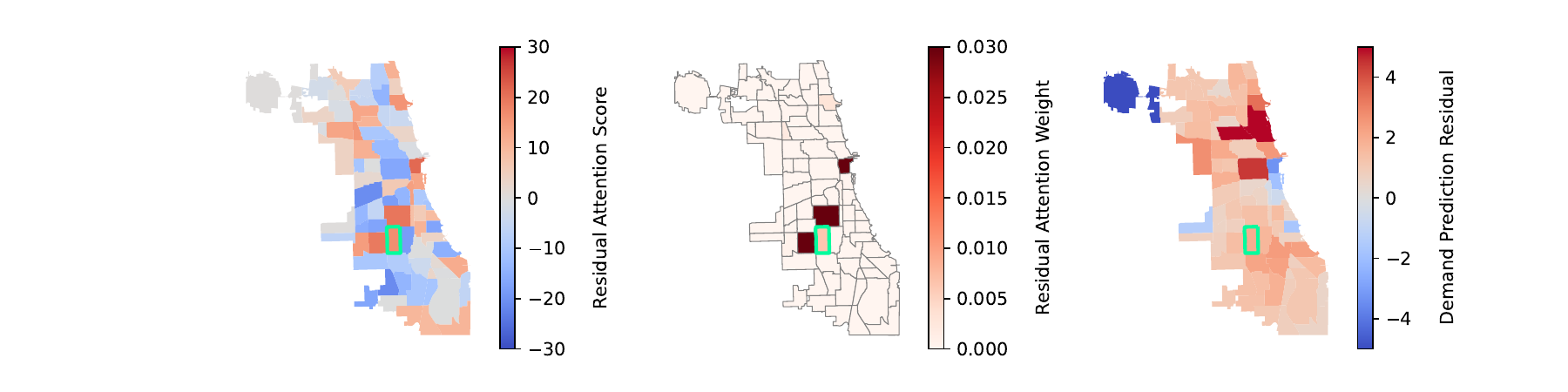}  
  \caption{Case Study 1: West Englewood. Attention map captures relationships with nearby areas.}
  \label{fig:attention1}
\end{subfigure}
\begin{subfigure}{\textwidth}
  \centering
  \includegraphics[width=.99\linewidth]{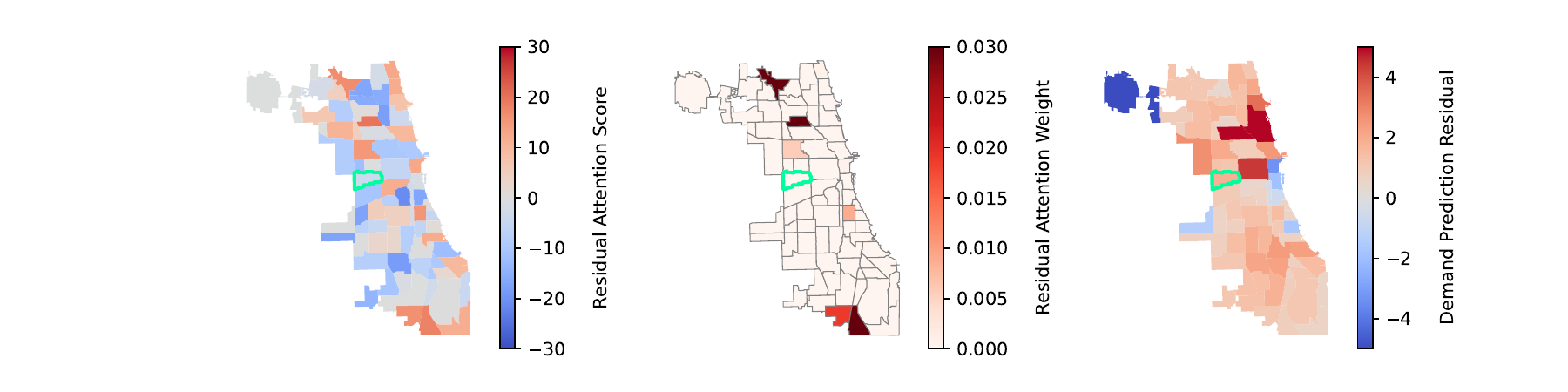}  
  \caption{Case Study 2: North Lawndale. Attention map captures relationships with faraway areas in both north and south directions.}
  \label{fig:attention2}
\end{subfigure}
\begin{subfigure}{\textwidth}
  \centering
  \includegraphics[width=.99\linewidth]{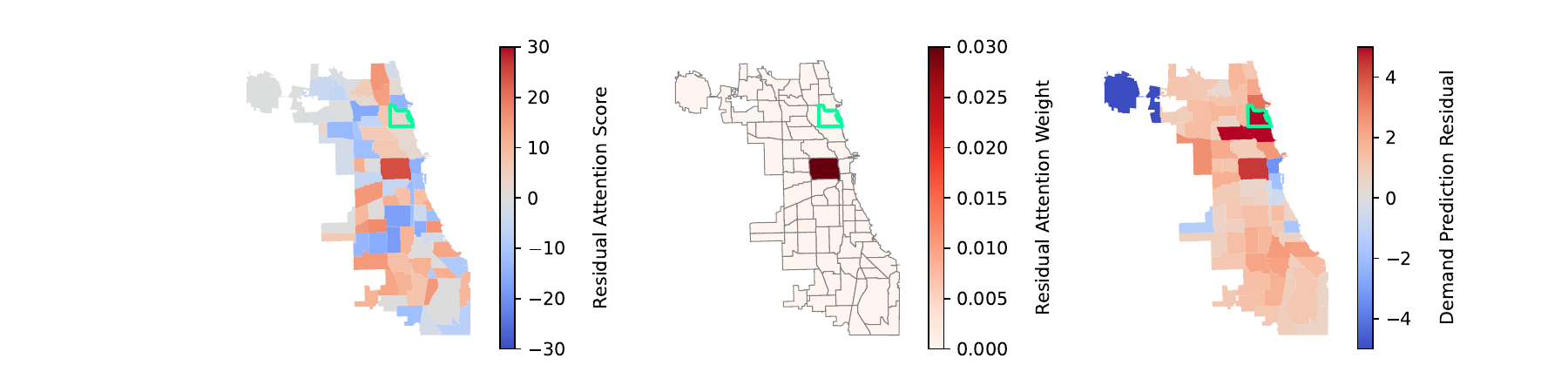}  
  \caption{Case Study 3: Lake View. Attention map captures relationships among under-prediction areas.}
  \label{fig:attention3}
\end{subfigure}
\begin{subfigure}{\textwidth}
  \centering
  \includegraphics[width=.99\linewidth]{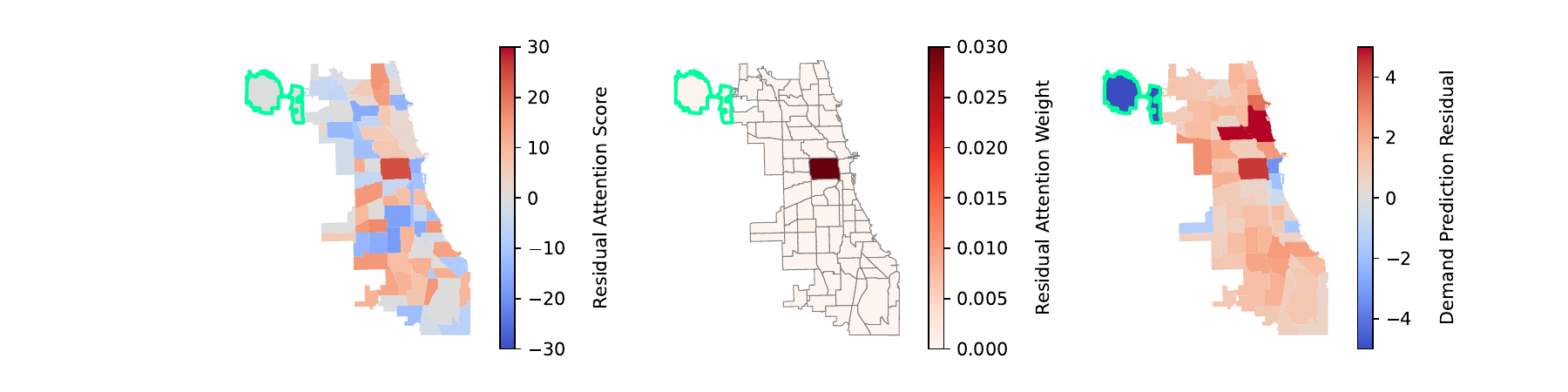}  
  \caption{Case Study 4: O'Hare. Attention map captures relationships between under- and over-prediction areas.}
  \label{fig:attention4}
\end{subfigure}
\caption{Attention Maps and Residuals in Chicago.}
\label{fig:attention_maps}
\end{figure}

To illustrate the influence of the Attention Mechanism in the RAA Block on the model, we plotted the attention score and attention map of the last epoch (explained in section \ref{sec:attention}) and compared them with the final prediction residuals for each region in Chicago, as shown in Fig \ref{fig:attention_maps}. 

We selected the results of four community areas predicted using the STGCN model with RAA block and the $D_s$ signed-aware residual variance term as regularization. 
The goal was to illustrate the characteristics of the attention maps for these community areas. 
For each case study, the focus community area is highlighted with a light green boundary line. 
The map on the left demonstrates the attention scores, representing the relationship between the given community area and the rest of the city. 
Regions shaded with darker red are associated with higher attention as learned from the residuals, whereas darker blue indicates the opposite.

To further highlight the most important attention regions in the city corresponding to the given community area, the middle map illustrates the attention weight calculated by applying the Softmax function to the attention scores. 
This visualization allows us to understand how the model focuses on different regions and how this focus is influenced by the residuals. 
The resulting attention maps provide insights into the spatial dependencies and the model’s adaptive learning process.

By comparing the attention scores and weights with the final prediction residuals, we can assess the effectiveness of the RAA Block in addressing spatial and demographic disparities. 
The attention mechanism helps the model to dynamically adjust its focus, leading to more equitable prediction results across different regions. 
This approach not only enhances the model’s accuracy but also ensures fairness in urban predictions, contributing to more balanced resource allocation and policy-making.

The four community areas for case studies include:

\textbf{Case Study 1: West Englewood. } In this scenario (Figure \ref{fig:attention1}), the attention related to the given community area concentrates in New City, Chicago Lawn, and the Loop areas. 
Notably, the first two regions are adjacent to the target West Englewood area. 
This demonstrates that the calculated attention weight prioritizes spatial proximity between community areas. 
This focus on nearby areas helps ensure that predictions are influenced more by spatially relevant information, potentially leading to more accurate and context-aware urban predictions. 

\textbf{Case Study 2: North Lawndale. } In this scenario (Figure \ref{fig:attention2}), the attentions corresponding to this community area focus on Forest Glen, Avondale, and Riverdale areas, which are spatially far from the target community area, North Lawndale. 
This demonstrates that the attention weight not only captures nearby entities but also emphasizes similar entities far away. 
The model's ability to highlight distant regions with comparable characteristics indicates that the attention mechanism effectively recognizes and leverages similarities in urban features, regardless of spatial distance. 

\textbf{Case Study 3: Lake View. } In this case (Figure \ref{fig:attention3}), the attentions focus on the Near West Side area, a busy central region of the city. 
Both Lake View and the Near West Side are underestimated, as shown in the demand prediction residual map. 
This illustrates that the attention weight is capable of linking regions with similar prediction performances. 
By identifying and connecting areas with analogous underestimation patterns, the model can better understand and correct systematic prediction biases. 
This aspect of the attention mechanism is particularly important in urban planning, where equitable resource distribution can significantly impact the quality of life for residents in different neighborhoods.

\textbf{Case Study 4: O'Hare. } In this case (Figure \ref{fig:attention4}), the attention also concentrates on the Near West Side area. However, while the O'Hare area is over-predicted, as shown in the demand prediction residual map, the Near West Side exhibits under-predictions. 
This reveals that the attention weight can connect regions with opposite behaviors, preventing the model from overemphasizing similarities. 
This balance helps in mitigating biases by ensuring that areas with different prediction errors are considered together, enhancing the model's ability to correct over-predictions and under-predictions effectively. 

% \hanyong{[Insert the maps of residuals prediction before and after RAA.]}

% \redfont{Introduce what is ablation study and why do we need it.}
\section{Discussion and conclusions}
\label{sec:conclusion}

Previous urban prediction work prevalently built on ST-GNNs focuses solely on accuracy, neglecting social impacts.
By focusing on residuals as indicators of fairness, we effectively highlight disparities in traditional ST-GNN outputs.
This study addresses spatial and demographic disparities in urban prediction tasks by developing an RAA Block and an equality-enhancing loss function. 
Our approach, integrated into existing ST-GNNs, dynamically adjusts spatial relationships during training, mitigating spatial disparities. 

Applied to urban prediction tasks in Chicago, our methodology demonstrates significant improvements both in fairness metrics and error metrics. 
This shows that reducing spatial disparities can also help mitigate demographic disparities.
Moreover, our approach reduces the local segregation of residuals and errors. 
Spatial analysis of residual distributions shows that models with RAA Blocks effectively reduced clustered prediction errors in central regions. 
Attention maps indicate the model's enhanced focus on relevant areas, improving prediction accuracy and fairness.
Through the comprehensive case study in Chicago, we demonstrate the effectiveness of our approach in mitigating prediction disparities for future equitable urban city management.

However, the RAA Block's effectiveness is model-dependent, and there is a trade-off between residual clustering, variance, and the balance of over- and under-predictions. 
Future work should optimize these factors to enhance performance. Although our approach does not rely on demographic data, incorporating such information when available could provide a more comprehensive understanding of fairness in urban predictions.

\section{Acknowledgement}
ChatGPT was utilized to assist in polishing the wording and beautifying the code. No other uses were conducted. We have thoroughly verified the accuracy, validity, and appropriateness of the content generated. No citations or literature reviews were performed by the language model. We acknowledge the limitations of language models in the manuscript, including potential bias, errors, and gaps in knowledge.

\bibliographystyle{plainnat}  
\bibliography{references}  %%% Remove comment to use the external .bib file (using bibtex).

\end{document}